\documentclass[letterpaper]{article} 
\usepackage{aaai24}
\usepackage{times}  
\usepackage{helvet}  
\usepackage{courier}  
\usepackage[hyphens]{url}  
\usepackage{graphicx} 
\usepackage{xcolor}
\usepackage{natbib}  
\usepackage{caption} 
\usepackage{algorithm}
\usepackage{algorithmic}

\usepackage{newfloat}
\usepackage{listings}
\DeclareCaptionStyle{ruled}{labelfont=normalfont,labelsep=colon,strut=off} 
\floatstyle{ruled}
\newfloat{listing}{tb}{lst}{}
\floatname{listing}{Listing}
\title{Evaluation and Enhancement of Semantic Grounding\\ in Large Vision-Language Models}
\author{
    Jiaying Lu\thanks{Work was done when Jiaying Lu was an intern at Mineral.}\textsuperscript{\ding{171}}\equalcontrib,
    Jinmeng Rao\textsuperscript{\textcolor{red}{\ding{169}}}\equalcontrib, 
    Kezhen Chen\textsuperscript{\color{red}\ding{169}}, 
    Xiaoyuan Guo\textsuperscript{\color{red}\ding{169}},
    Yawen Zhang\textsuperscript{\color{red}\ding{169}}, 
    Baochen Sun\textsuperscript{\color{red}\ding{169}}, 
    Carl Yang\textsuperscript{\ding{171}}, 
    Jie Yang\textsuperscript{\color{red}\ding{169}}
}
\affiliations{
    \textsuperscript{\color{red}\ding{169}}Mineral,
    \textsuperscript{\ding{171}}Emory University\\
    \textsuperscript{\color{red}\ding{169}}\{jinmengrao, kezhenchen, xiaoyuanguo, yawenz, baochens, yangjie\}@mineral.ai\\
    \textsuperscript{\ding{171}}\{jiaying.lu, j.carlyang\}@emory.edu
}

\usepackage{bibentry}

\usepackage{caption}
\usepackage{subcaption}
\usepackage{graphicx}
\usepackage{amsmath}
\usepackage{amssymb}
\usepackage{booktabs}
\usepackage{enumitem}
\usepackage{multirow}
\usepackage{xspace}
\usepackage{tabularx}
\usepackage{arydshln}
\usepackage{xcolor}
\usepackage{array}
\usepackage[title]{appendix}
\usepackage{pifont}
\usepackage{pgfplots}
\pgfplotsset{compat=1.16}
\usepackage{tkz-kiviat}
\usetikzlibrary{arrows}

\newcommand{\header}[1]{\noindent \textbf{#1}}

\newcommand{\eval}[0]{\textsc{Msg-Mcq}\xspace}
\newcommand{\imp}[1]{\textit{\footnotesize ($\uparrow$#1)}}
\newcommand{\dtrt}[1]{\textit{\footnotesize ($\downarrow$#1)}}
\newtheorem{definition}{Definition}
\newtheorem{example}{Example}
\definecolor{RadarC1}{HTML}{7fc97f}
\definecolor{RadarC2}{HTML}{beaed4}
\definecolor{RadarC3}{HTML}{fdc086}
\definecolor{RadarC4}{HTML}{ffff99}
\definecolor{RadarC5}{HTML}{386cb0}
\definecolor{RadarC6}{HTML}{f0027f}
\definecolor{RadarC7}{HTML}{bf5b17}

\usepackage[capitalize]{cleveref}
\crefname{section}{Sec.}{Secs.}
\Crefname{section}{Section}{Sections}
\Crefname{table}{Table}{Tables}
\crefname{table}{Tab.}{Tabs.}

\begin{document}

\maketitle

\begin{abstract} 
Large Vision-Language Models (LVLMs) offer remarkable benefits for a variety of vision-language tasks. However, a challenge hindering their application in real-world scenarios, particularly regarding safety, robustness, and reliability, is their constrained \emph{semantic grounding} ability, which pertains to connecting language to the physical-world entities or concepts referenced in images. Therefore, a crucial need arises for a comprehensive study to assess the semantic grounding ability of widely used LVLMs. Despite the significance, sufficient investigation in this direction is currently lacking. Our work bridges this gap by designing a pipeline for generating large-scale evaluation datasets covering fine-grained semantic information, such as color, number, material, \emph{etc.}, along with a thorough assessment of seven popular LVLMs' semantic grounding ability. Results highlight prevalent \emph{misgrounding} across various aspects and degrees. To address this issue, we propose a data-centric enhancement method that aims to improve LVLMs' semantic grounding ability through multimodal instruction tuning on fine-grained conversations. Experiments on enhanced LVLMs demonstrate notable improvements in addressing misgrounding issues.
\end{abstract}



\section{Introduction}
\label{sec:intro}

Large Vision-Language Models (LVLMs)~\cite{zhao2023chatbridge,tang2023any,openai2023gpt4,inner_monologue} expand the powerful large language models~\cite{ouyang2022instructGPT,touvron2023llama2,ling2023beyond} to versatile general-purpose vision-language understanding and generation interfaces. This is achieved through the integration of the vision encoder and the autoregressive large language model~\cite{liu2023LLaVA,li2023otter,luo2023LaVIN}.
While demonstrating promising performance in solving various vision-language benchmarks, comprehensive examination and analysis are desired before deploying LVLMs into real-world critical-sensitive applications. Recent studies reveal that many LVLMs still suffer from text-image misalignment~\cite{yarom2023SeeTRUE}, adversary perturbed input~\cite{zhao2023adversarial}, and object hallucination~\cite{li2023POPE} even in some seemingly simple cases. 
In this paper, we specifically focus on the evaluation of the under-explored semantic grounding ability in LVLMs.

Semantic grounding (\emph{i.e.} the capability to connect words to the physical-world entities or concepts they refer to)~\cite{yun2021grouding,li2022grounded} is critical to the safe, robust, and reliable development of LVLMs. 
Although identifying ``a Siberian tiger'' as ``a tabby cat'' in the user-shared image under a social chatbot setting seems harmless, stakes escalate significantly when an LVLM-aided disease diagnosis assistant interprets instruction to analyze the patient's ``left lung'' as ``right lung''.
In this study, we comprehensively evaluate the semantic grounding ability of existing LVLMs through our proposed \emph{evaluation suite}. Specifically, we automatically generate $9,000$ vision-language test samples, exploring semantic grounding proficiency across four formats and addressing six types of grounding targets. Seven state-of-the-art LVLMs are evaluated using the $9,000$ test samples, and the experimental results reveal that most of them exhibit semantic grounding deficiency across various aspects and degrees. 
To facilitate the scalability of test sample generation and the interpretability of evaluation metrics, we propose to adopt multiple-choice questions as the test format. Further technical details are presented in the \emph{Evaluation of Semantic Grounding} Section.

Moreover, we introduce a data-centric \emph{enhancement method} designed to enhance the semantic grounding capabilities of LVLMs. Diverging from classical model-centric approaches~\cite{zha2023dcai} that emphasize advancements in model architecture, our approach focuses on curating a substantial volume of diverse multimodal, multi-round conversation data that can be leveraged by any LVLM. In total, we curate $180,000$ fine-grained instructional instances for the purpose of semantic grounding enhancement. Experimental results with LVLMs fine-tuned on our enhancement data consistently demonstrate improvements in multimodal semantic grounding.
\section{Related Work}
\label{sec:related_work}
\header{Trustworthiness Evaluation of LVLMs}.
Many works leverage existing vision-language datasets to derive trustworthiness evaluation benchmarks for LVLMs. MME~\cite{fu2023mme} consists of 14 subtasks based on public images with manually constructed annotations, which measure both perception and cognition abilities of LVLMs in the form of Yes-or-No question answering. The LAMM benchmark~\cite{yin2023lamm} covers nine common 2D image tasks and three common point cloud tasks with specifically curated inference instruction. Other similar benchmarks include LVLM-eHub~\cite{xu2023lvlmEhub}, MM-Vet~\cite{yu2023mmvet}, and MMBench~\cite{liu2023mmbench}, etc. 
There also exist benchmarks focusing on evaluating specific properties of LVLMs. POPE~\cite{li2023POPE} focuses on evaluating object hallucination by asking Yes-or-No questions regarding the object existence of input images. M-HalDetect~\cite{gunjal2023MHalDetect} proposes the hallucination task as sentence-level classification, using human-annotated labels. \cite{zhao2023adversarial} proposes evaluating the robustness of LVLMs by adding adversarial noise into input images.
Our work also provides a valuable resource to serve as a comprehensive trusworthiness benchmark for LVLMs, from a novel perspective focusing on semantic grounding. Moreover, we provide a generic framework to enable researchers to conveniently create test samples for evaluating LVLMs.

\header{Methods for Building Safe, Robust, and Reliable LVLMs}.
There are two streams of approaches to improve the safeness, robustness, and responsibilities of LLMs and LVLMs~\cite{ji2023hallucinationSurvey}: model-centric approaches and data-centric approaches. 
Model-centric approaches often focus on the model advancements, which involve (1) designing robust training paradigms~\cite{berg2022debias,dong2023co}, (2) robust inference~\cite{wang2022self,zhang2023MCoT}, (3) refining generated response~\cite{madaan2023selfRefine}, \textit{etc}.
Data-centric approaches~\cite{zha2023dcai,bai2024efficiency}, on the other hand, focus on ensuring data quality and reliability, which often involve (1) faithful training data development such as data collection~\cite{liu2023LLaVA,lu2023MuG} and data cleaning~\cite{northcutt2021confident,monarch2021human}, (2) inference stage data augmentation such as retrieving supporting knowledge~\cite{chen2022murag,cui2023survey}.
Our framework follows the data-centric idea. Instead of modifying the model structure, our work aims to improve the semantic grounding ability of LVLMs via multimodal instruction tuning, which has proved to be a generic and efficient approach for improving LVLMs.
\section{Preliminaries}
\label{sec:prelim}

\label{ssec:LVLM}
\begin{figure}[htbp!]
    \centering
    \includegraphics[width=0.75\linewidth]{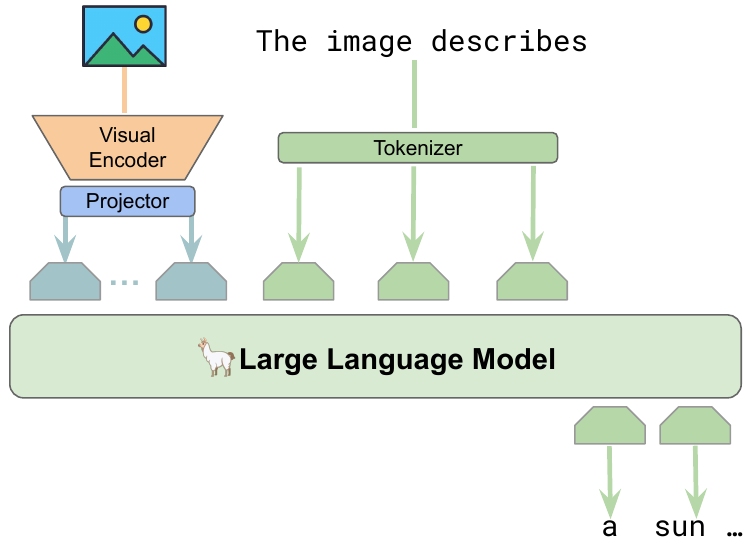}
    \caption{One representative architecture of existing LVLMs. \textcolor{cyan}{Cyan} hexagons denote visual embeddings, and \textcolor{green}{green} hexagons denote textual embeddings.}
    \label{fig:mllm_arch}
\end{figure}

We illustrate one representative architecture of LVLMs in Figure~\ref{fig:mllm_arch}, which typically consists of the following components: 
\begin{enumerate}
    \item a visual encoder that extracts features from the input image;
    \item a projector that projects visual features to the language embedding space;
    \item a tokenizer that tokenizes textual input into textual tokens and maps them into language embedding space;
    \item a decoder-only LLM (e.g., LLaMA) that generates textual responses based on the multimodal inputs. 
\end{enumerate}
The overall output generation process of a LVLM $\mathcal{F}_{\Theta}$ can be formally described by 
\begin{equation}
\mathbf{Y}=\mathcal{F}_{\Theta}(\mathbf{X}^v,\mathbf{X}^q),
\end{equation}
where $\Theta$ are parameters of the LVLM, $\mathbf{X}^v$ denotes the visual input, $\mathbf{X}^q$ denotes the textual input, and $\mathbf{Y}$ denotes the generated output sequence.

Specifically, given a visual input $\mathbf{X}^v$, a visual encoder $f_{\phi}$ firstly extracts visual features by
\begin{equation}
    \mathbf{Z}^v=f_{\phi}(\mathbf{X}^v),
\end{equation}
where $\mathbf{Z}^v \in \mathbb{R}^{l^v\times d}$, and $\phi$ are parameters of $f_{\phi}$ typically frozen in LVLMs training. Here $l^v$ denotes the length of visual tokens, and $d$ is the dimension of visual tokens.
Regarding the visual encoder $f_{\phi}$, most existing LVLMs employ ViT-based structures~\cite{dosovitskiy2020ViT} and select certain layers outputs to construct a certain length of visual tokens. For instance, LLaVA~\cite{liu2023LLaVA} utilizes grid features before and after the last Transformer layer of ViT, while LaVIN~\cite{luo2023LaVIN} utilizes the $[CLS]$ embeddings from every fourth layer of ViT. 

A trainable projector, denoted as $g_{\omega}$, is then applied to convert $\mathbf{Z}^v$ into language embedding space, which can be defined by 
\begin{equation}
\label{eq:projector}
    \mathbf{H}^v=g_{\omega}(\mathbf{Z}^v)=g_{\omega}(f_{\phi}(\mathbf{X}^v)),
\end{equation}
where $\mathbf{H}^v \in \mathbb{R}^{l^v\times h}$, and $\omega$ are trainable parameters of $g_\omega$. $h$ denotes the dimension of language embedding space.
In practise, an efficient implementation of $g_{\omega}$ can be a projection matrix $\mathbf{W}\in \mathbb{R}^{d\times h}$~\cite{liu2023LLaVA}, thus $\mathbf{H}^v=\mathbf{Z}^v\mathbf{W}$. More sophisticated projectors are also proposed, such as Q-former~\cite{li2023blip2}.

Given the input textual query $\mathbf{X}^q$, a tokenizer $k_{\psi}$ is employed to tokenize and map them into ``textual tokens'' by:
\begin{equation}
\label{eq:tokenizer}
    \mathbf{H}^q=k_{\psi}(\mathbf{X}^q),
\end{equation}
where $\mathbf{H}^q \in \mathbb{R}^{l^q\times h}$, and $\psi$ are trainable parameters of $k_{\psi}$. It is worth noting that $\mathbf{H}^q$ and projected $\mathbf{H}^v$ have the same dimensionality $\mathbb{R}^h$ as the input of the large language model $p_{\theta}$. 
Similarly, the length of textual tokens varies depending on the choice of tokenizer, as words are chunked into subwords. Moreover, special tokens $[s], [/s]$ can be added to indicate the span of visual tokens and textual tokens.
Therefore, we denote the process of LVLM to prepare the multimodal input of its LLM component as:
\begin{equation}
    \mathbf{H} = [\mathbf{H}^v,\mathbf{H}^q] \in \mathbb{R}^{(l^q+l^v)\times h}.
\end{equation}

Once the multimodal input features are processed, the LLM component of an LVLM is responsible for generating responses. The LLM component is essentially a probabilistic auto-regressive model $p_{\theta}$ with trainable parameters $\theta$ that can predict the next token $y_t$ step by step based on the input $\mathbf{H}$ and tokens predicted so far $\mathbf{Y}_{0:t-1}$. The process can be formulated by:
\begin{equation}
\label{eq:LLM_infer}
    Pr(\mathbf{Y}_{0:\tau}|\mathbf{H}) = \prod_{t=1}^{\tau} p_{\theta}(y_t|\mathbf{H},\mathbf{Y}_{0:t-1}).
\end{equation}
where $Pr$ denotes the probability of generating the output sequence $\mathbf{Y}_{0:\tau}$.

\section{Evaluation of Semantic Grounding}
\label{sec:eval}

\begin{figure*}[!ht]
    \centering
    \includegraphics[width=\linewidth]{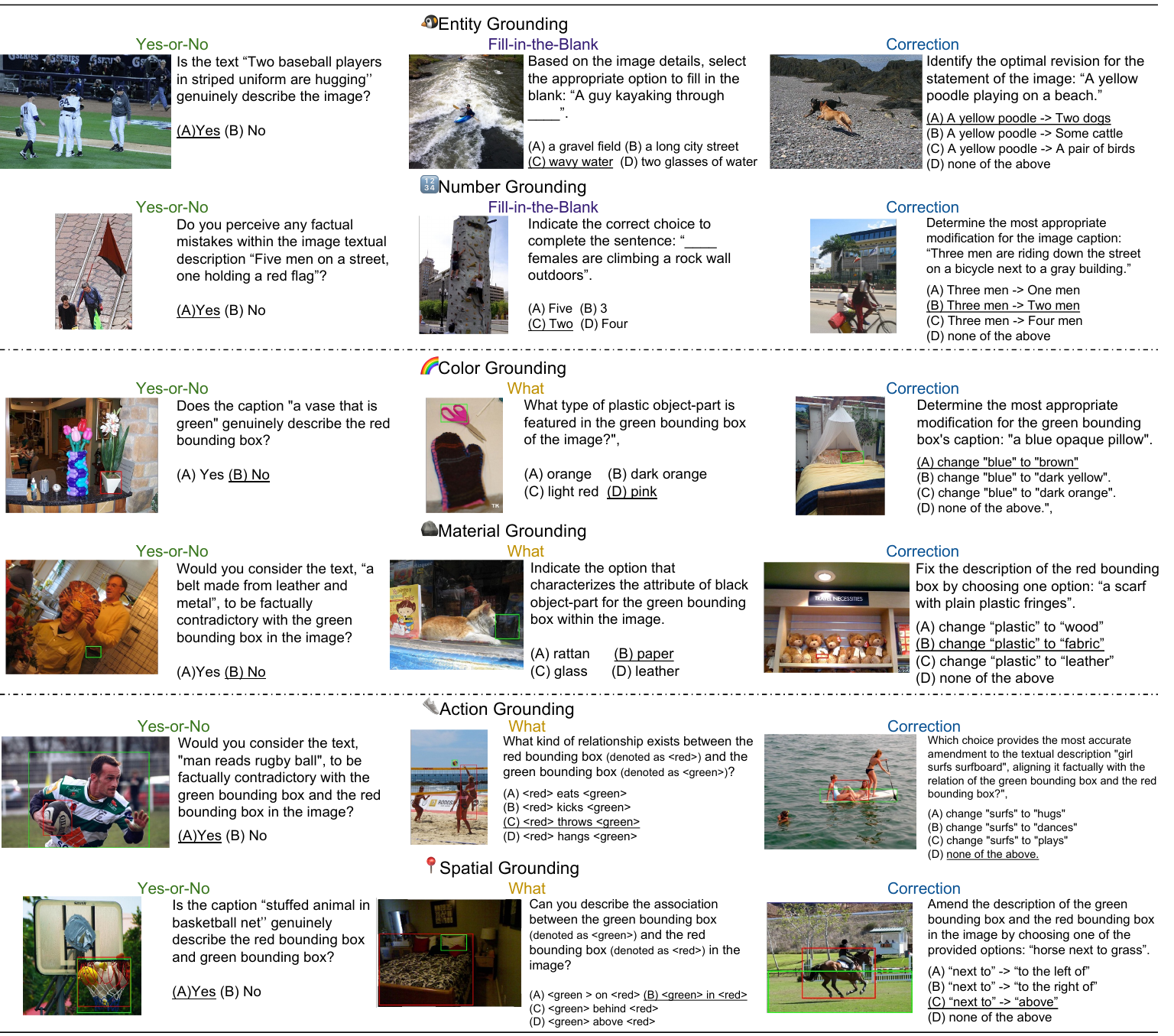}
    \caption{Instances of our \eval, which covers four types of MCQs and six types of grounding targets.}
    \label{fig:benchmark_instance}
\end{figure*}

A wide range of forms can be used for the evaluation of semantic grounding in LVLMs, and each form has its own advantages and disadvantages. Free-form questions~\cite{yarom2023SeeTRUE} are easy to design but require resource-intensive human evaluations and are difficult to score consistently. Similarity-based evaluation uses less resources, but is heavily reliant on high-quality ground truth responses and the bias of similarity metrics.
Yes-or-No questions~\cite{fu2023mme} are non-ambiguous and easier to evaluate. However, they may be too easy and cannot capture all aspects of semantic grounding in LVLMs.

We propose \textbf{\eval} for the novel \textbf{M}ultimodal \textbf{S}emantic \textbf{G}rounding evaluation based on \textbf{M}ultiple-\textbf{C}hoice \textbf{Q}uestions (MCQs)~\cite{lu2022good,lu2022scienceQA}.
This form presents a question along with a set of predetermined choices, allowing respondents to select the option they believe to be correct. The MCQs facilitate efficient grading and analysis of responses, and the difficulty level can be controlled by adjusting the number and the feasibility of distractor choices ~\cite{lu2022good,lu2022scienceQA}. Moreover, the Yes-or-No form can be regarded as a special case in MCQs where the choices are ``(A)Yes. (B) No.''. We use \emph{accuracy} as the evaluation metric for MCQs.
Following the notation in the \emph{Preliminaries} section, we extend the textual input question $\mathbf{X}^q$ into two parts: the question body $\mathbf{X}^{qb}$ and the multiple choices $\mathbf{X}^{qc}$. Therefore, we formally define the task of \eval:

\begin{definition}[\eval]
Given an input image $\mathbf{X}^{v}$, input question body $\mathbf{X}^{qb}$ , and $K$ input choices $\mathbf{X}^{qc}=\{\mathbf{C}^1,\mathbf{C}^2,\dots,\mathbf{C}^K \}$, \eval expects a LVLM $\mathcal{F}_{\Theta}$ to select the correct choice $\mathbf{C}^i$ where $1\leq i \leq K$ from the input choices $\mathbf{X}^{qc}$.
\end{definition}
In the scope of LVLM, the selection action is typically determined by the generated sequence of LVLM. Depending on the specific implementation of LVLMs, the output can be as simple as a choice indicator like ``C'', or a comprehensive response that includes both the selected choice and an explanation of selecting it such as ``The correct choice is A, because $\dots$''.

\subsection{\eval Generation Pipeline}
In this work, we propose an efficient automated MCQ generation method to derive our \eval.
Figure~\ref{fig:benchmark_instance} provides several automatically generated evaluation examples. 
Specifically, we pose four specific kinds of MCQs: \emph{Yes-or-No, Fill-in-the-Blank, What, Correction} in different columns in Figure~\ref{fig:benchmark_instance}. Orthogonal to the specific kinds of MCQs, the evaluation focuses on different targets of semantic grounding that indicate the particular deficiency of evaluated LVLMs. In this work, we include six targets of semantic grounding: \emph{Entity, Number, Color, Material, Action, and Spatial}. These target categories are shown in different rows in Figure~\ref{fig:benchmark_instance}. It is worth noting that the evaluation module can be easily extended to more types and targets in the future.

All these MCQs are generated from ground-truth source data through a four-step generation pipeline. In \textbf{step 1}, we curate a pool of question templates with placeholders for each kind of MCQ focusing on one specific target of semantic grounding, according to the source data we have. In \textbf{step 2}, we randomly sample one question template from the pool. As an illustration, a question template of \emph{What} question focusing on color grounding is: 
\begin{example}[What question]
What color of [obj-attr] object is featured in [bbox-color] bounding box of the image? \\
(A) [distractor\#1] \qquad (B) [ground-truth] \\ 
(C) [distractor\#2] \qquad (D) [distractor\#3]. 
\end{example}

As can be seen, the sampled question template contains several placeholders, and the ground-truth is randomly placed into the choice (A), (B), (C), or (D).
In \textbf{step 3}, we use source data to fill in these placeholders. For example, ``\emph{[obj-attr], [ground-truth]}'' can be filled using the gold annotation in the source data. Regarding the ``\emph{[bbox-color]}'', we first use gold bounding box coordinates referring to the querying object to draw a box in the original image using a random color (green or red), then replace ``\emph{[bbox-color]}'' with the name of that color.
In \textbf{step 4}, we generate the distractors of ground truth based on the multimodal input information. This is one of the most critical steps in the evaluation pipeline, since distractors determine the difficulty level of MCQs. Distractors should not be semantically equivalent to ground truth, but they should be plausible enough to serve as an answer candidate to the question. Various distractor generation methods~\cite{lu2022good} can be used, such as manual generation, thesaurus-based generation, and end-to-end generation. In this work, we utilize thesaurus-based generation with post-human verification. According to the type of ground truth (entity, number, color, \textit{etc.}), we randomly sample $15$ distractors. For entity grounding, we use the sentence transformer model~\cite{reimers-2019-sentence-bert} to filter out candidates that have overly high similarity scores to the gold answer. For other targets of grounding, the thesaurus is guaranteed to contain semantically different antonyms of the gold answer.


We introduce the necessary tweak of the overall pipeline for each specific kind of MCQ, and explain the purpose of why we need them as below:

\begin{itemize}
    \item \textbf{Yes-or-No:} \emph{Can a model identify whether a textual description is appropriate for a given image?} For each Yes-or-No MCQ, two choices are given, as shown in the first column of Figure~\ref{fig:benchmark_instance}. The correct textual descriptions are either directly taken from the source data (image captioning datasets), or obtained using some sentence templates (object detection). The wrong descriptions are generated by negative replacement on correct descriptions, where we change specific textual spans for corresponding semantic grounding target (\textit{e.g.} sampling ``yellow'' to replace the original ``blue'' in color grounding).
    \item \textbf{Fill-in-the-Blank.}  \emph{Can a model infer the missing pieces of information regarding multimodal input?} For each Fill-in-the-Blank MCQ, four choices are given where only one choice is correct. The second column of the first two rows of Figure~\ref{fig:benchmark_instance} gives real examples of such type MCQs. A text span that contains the concept of interest (\textit{e.g.} entity, number) is blanked, and the span is used as one choice. Three distracting choices are generated using negative sampling.
    \item \textbf{What.} \emph{Can a model recognize and identify an object or attribute that is specified in the multimodal input?} For each What MCQ, four choices are given where only one choice is correct. The second column of the last four rows of Figure~\ref{fig:benchmark_instance} gives real examples of such type MCQs. Similarly, three distracting choices are generated using negative sampling.
    \item \textbf{Correction} \emph{Can a model identify the inconsistency across different modalities and propose an appropriate correction?} For each Correction MCQ, four choices are given where only one choice is correct. The last column of Figure~\ref{fig:benchmark_instance} provides real examples of such type MCQs. The choice ``(D) none of the above'' is always included in each Correction MCQ, indicating the scenario where the original description is grounded. Correction MCQs are challenging, as they require models to imagine whether the corrected text is consistent with the image.
\end{itemize}
\section{Enhancement of Semantic Grounding}
Our solution for enhancing the semantic grounding of LVLMs is a data-centric approach~\cite{zha2023dcai} to systematically and algorithmically generate the instruction-tuning dataset to feed LVLMs. The emerging success of transformer model architectures in both large language models~\cite{ouyang2022instructGPT,openai2023gpt4,touvron2023llama2} and large vision-language models~\cite{zhu2023minigpt,awadalla2023openflamingo} advocates a fundamental shift from the model-centric AI to data-centric AI. Instead of focusing on designing specific model architectures for a particular downstream task, we curate a substantial (180K instances) of new multimodal multi-round conversational instruction data for LVLMs to be further improved upon. 

\subsection{Instruction Tuning Data Generation}
\begin{table}[!h]
\centering
\begin{tabular}{lp{0.8\linewidth}}
\toprule
Image & \includegraphics[width=0.85\linewidth]{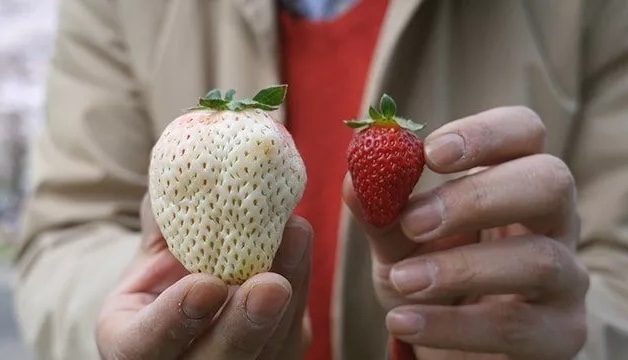}\\
\\
User & What colors are the strawberries the man is holding?\\
LVLM & The man is holding one white strawberry and one red strawberry.\\
User & Do you think the white strawberry is to the right of the red strawberry?\\
LVLM & No, the white strawberry appears to be on the left of the red strawberry in the image.\\
\bottomrule
\end{tabular}
\caption{Toy Example of Multimodal Multi-Round Instruction-Tuning Data}
\label{tab:instruction_data}
\end{table}

For the instruction data generation, we first curate a set of fine-grained conversations of multimodal inputs that help LVLMs interpret the information in the image step by step and eventually answer the question. 
Table~\ref{tab:instruction_data} provides a toy example of such instruction data.
All the conversations are provided as multi-step chain-of-thought~\cite{zhang2023MCoT} instructions to simulate the human reasoning process and guide LVLMs to pay attention to the right information at each step. Specifically, we design three types of instructions:

\begin{itemize}[noitemsep]
    \item \textbf{Multi-Round Conversation}. Given an image, we provide a multi-round conversation between users and models. The user asks a few questions about some fine-grained content in the image and the LVLM gives factual and precise answers based on the observation.
    \item \textbf{Vision-Prompted Recognition}. Given an image, we draw visual prompts (e.g., bounding boxes) on the image and ask an LVLM to tell the names, attributes, or relations of the objects indicated by visual prompts. The LVLM needs to learn how to follow the visual prompts to first localize the indicated objects and then recognize detailed attributes or relations of them.
    \item \textbf{Fact Checking}. Given an image and a statement of some facts in the image, we ask the LVLM if and why a given statement is factually consistent. In the instructions, we provide reference answers to guide the LVLM to localize main objects, recognize their attributes and relations, and determine if there are any factual misalignments. Eventually, the LVLM determines whether it is factual.
\end{itemize}
All three types of instructions essentially consist of two fundamental components: the question and the response. We manually curate diverse templates for both questions and responses. In a similar fashion, the response templates include placeholders to be filled with ground-truth data. Once these placeholders are populated, we employ chatGPT to rephrase these instructions, enhancing their diversity.

\subsection{Instruction Tuning}
After the fine-grained conversations are generated, we conduct instruction tuning~\cite{liu2023LLaVA,ouyang2022instructGPT} on LVLMs to enhance semantic grounding. 
Following the notation we introduce in the \emph{Preliminary section}, the inference process of LVLM can be described as: 
\begin{equation}
    \mathbf{\hat{Y}} = \mathcal{F}_{\Theta}(\mathbf{X}^v, \mathbf{X}^q),
\end{equation}
where $\mathcal{F}_{\Theta}$ denotes the LVLM with trainable parameters $\Theta$, $\mathbf{X}^v$ denotes the visual inputs, $\mathbf{X}^q$ denotes the textual inputs, and $\mathbf{\hat{Y}}$ denotes the textual response. It is worth noting that the trainable parameters $\Theta$ of $\mathcal{F}$ actually contains $\Theta=\{\phi, \omega, \psi, \theta \}$.
We denote the $n$-th training sample $(\mathbf{X}^v_{(n)}, \mathbf{X}^q_{(n)}, \mathbf{Y}^r_{(n)})$ from our generated instruction tuning data, where they represent the visual input, textual query input, and the ground truth response, respectively. The loss function of LVLM is 
\begin{equation}
     \mathcal{L}(\Theta) = -\sum_{n=1}^{N} log\, Pr(\mathbf{Y}^r_{(n)}|\mathbf{X}^v_{(n)},\mathbf{X}^q_{(n)}),
\end{equation}
where $Pr$ denotes the probability of generating the output sequence $\mathbf{Y}^r_{(n)}$ from input $(\mathbf{X}^v_{(n)},\mathbf{X}^q_{(n)})$ using equation~\eqref{eq:LLM_infer}.

\section{Experiments and Analysis}
\begin{table*}[h!]
    \centering
    \begin{tabular}{cccccc}
    \toprule
    {\footnotesize Grounding Target} & Q Type & \#IMGs & \#Qs & \#As & Data Source\\
    \midrule
    Entity & {\footnotesize YoN, FiB, Corr}& 1,339 & 1,500 & 628& Flickr30K \\
    Number & {\footnotesize YoN, FiB, Corr}& 977 &  1,500& 389& Flickr30K \\
    Color & {\footnotesize YoN, What, Corr}& 1,500 &  1,500 & 153& PACO \\
    Material & {\footnotesize YoN, What, Corr}& 1,500 & 1,500 & 68& PACO \\
    Action & {\footnotesize YoN, What, Corr}& 1,498 & 1,500 & 322 & OpenImage-V7 \\
    Spatial &  {\footnotesize YoN, What, Corr}& 1,500 & 1,500 & 144& SpatialSense \\
    \hline
    Overall& {\footnotesize YoN, FiB, What, Corr}& 8,119& 9,000& 1,683&\\
    \bottomrule
    \end{tabular}
    \caption{Overview of MCQs generated by \eval. \#IMGs denotes unique images, \#Qs denotes unique questions, and \#A denotes unique answers.}
    \label{tab:benchmark_overview}
\end{table*}

\subsection{Evaluation Examples}
Table~\ref{tab:benchmark_overview} gives an overview of key statistics of test multiple-choice questions generated by \eval. 
We collect six subsets for each semantic grounding type, together with 9K MCQs, 8.1K images, and 1.7K unique answers. 
These MCQs are built from four public datasets: Flickr30K~\cite{flickr30k}, PACO~\cite{ramanathan2023paco}, OpenImage-V7~\cite{OpenImages}, SpatialSense~\cite{yang2019spatialsense}. 
Flickr30K is an image captioning dataset that includes images obtained from Flickr and each image is provided with five manually annotated captions. PACO, OpenImage-V7 and SpatialSense are object detection datasets that contain fine-grained object/object-part bounding boxes, categories, and attribute annotations. Furthermore, OpenImage-V7 and SpatialSense provide relational annotations between two objects within one image.
We curate these testing MCQs by our proposed \eval. Also, we balance the answer distribution and the concept of semantic grounding target for promising evaluation.

\newcommand\ColorBox[1]{\textcolor{#1}{\rule{2ex}{2ex}}}
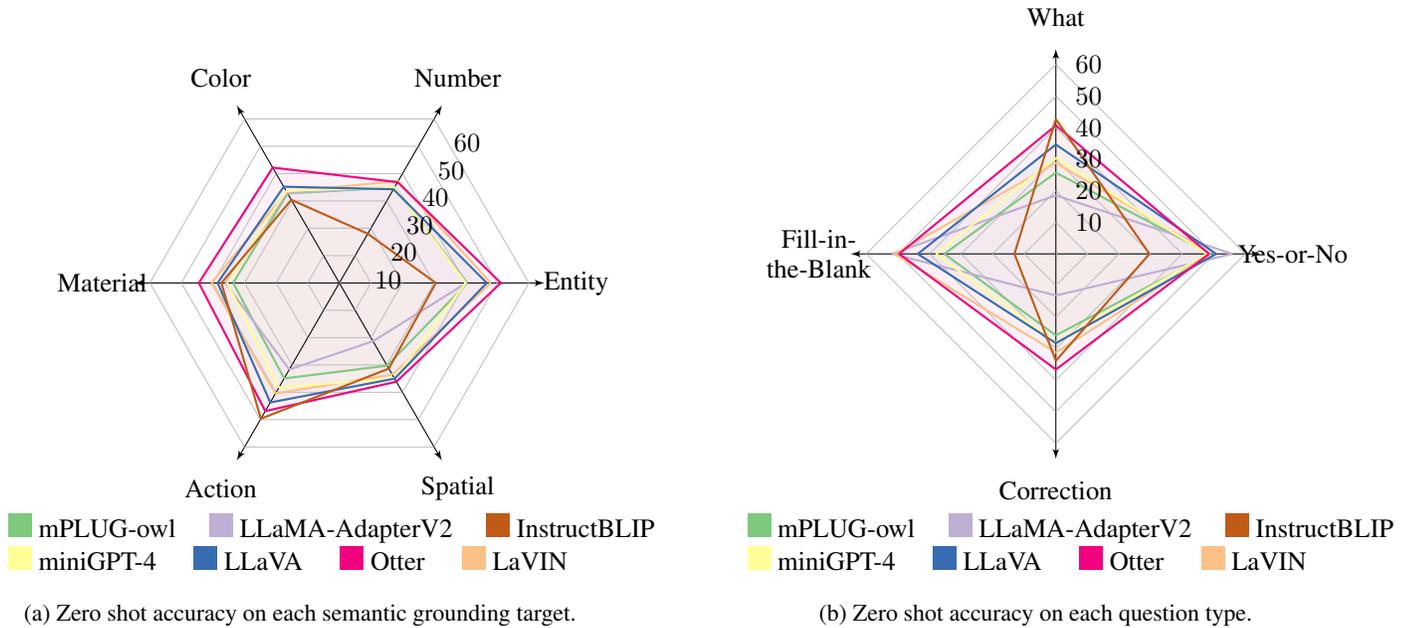
\begin{figure*}[!h]
\centering
  \begin{subfigure}[b]{0.45\textwidth}
  \begin{tikzpicture}
\tkzKiviatDiagram[
        scale=0.42,
        radial  = 6,
        gap     = 1,  
        lattice = 6]{Entity, Number, Color, Material, Action, Spatial}
\tkzKiviatLine[thick,color=RadarC1,mark=none,
               fill=RadarC1!20,opacity=.5](4.053, 3.493, 3.273, 3.360, 3.493, 3.027)
\tkzKiviatLine[thick,color=RadarC2,mark=none,
              fill=RadarC2!20,opacity=.5](4.020, 3.520, 3.293, 3.613, 3.167, 2.133)
\tkzKiviatLine[thick,color=RadarC3,mark=none,
              fill=RadarC3!20,opacity=.5](4.793, 3.727, 3.333, 4.047, 4.046, 3.353)
\tkzKiviatLine[thick,color=RadarC4,mark=none,
              fill=RadarC4!20,opacity=.5](4.000, 3.507, 3.367, 3.553, 3.827, 3.493)
\tkzKiviatLine[thick,color=RadarC5,mark=none,
              fill=RadarC5!20,opacity=.5](4.667, 3.427, 3.520, 3.853, 4.373, 3.500)
\tkzKiviatLine[thick,color=RadarC6,mark=none,
              fill=RadarC6!10,opacity=.5](5.113, 3.687, 4.220, 4.447, 4.687, 3.613)        
\tkzKiviatLine[thick,color=RadarC7,mark=none,
              fill=RadarC7!10,opacity=.5](3.053, 1.800, 3.040, 3.733, 4.967, 3.133)   
\tkzKiviatGrad[prefix=,unity=10](1)  
\node[shift={(0cm,-0.5cm)}] at (current bounding box.south) 
{
\begin{tabular}{@{}ll@{}}
\ColorBox{RadarC1} mPLUG-owl \hspace*{0.27cm} \ColorBox{RadarC2} LLaMA-AdapterV2 \hspace*{0.27cm} \ColorBox{RadarC7} InstructBLIP\\
\ColorBox{RadarC4} miniGPT-4 \hspace*{0.32cm} \ColorBox{RadarC5} LLaVA \hspace*{0.32cm} \ColorBox{RadarC6} Otter \hspace*{0.32cm} \ColorBox{RadarC3} LaVIN \\
\end{tabular}
};
\end{tikzpicture}
  \vspace{-0.4cm}
  \caption{Zero shot accuracy on each semantic grounding target.}
  \label{sfig:radar_agnosia}
  \end{subfigure}
\hfill
  \begin{subfigure}[b]{0.45\textwidth}
  \begin{tikzpicture}
\tkzKiviatDiagram[
        scale=0.42,
        radial  = 4,
        gap     = 1,  
        lattice = 6]{Yes-or-No, What, Fill-in-the-Blank, Correction}
\tkzKiviatLine[thick,color=RadarC1,mark=none,
               fill=RadarC1!20,opacity=.5](4.877, 2.575, 3.530, 2.580)
\tkzKiviatLine[thick,color=RadarC2,mark=none,
               fill=RadarC2!20,opacity=.5](5.590, 1.855, 5.190, 1.317)
\tkzKiviatLine[thick,color=RadarC3,mark=none,
               fill=RadarC3!20,opacity=.5](4.900, 2.885, 5.110, 3.123)
\tkzKiviatLine[thick,color=RadarC4,mark=none,
              fill=RadarC4!20,opacity=.5](4.777, 3.045, 3.790, 2.803)
\tkzKiviatLine[thick,color=RadarC5,mark=none,
              fill=RadarC5!20,opacity=.5](5.077, 3.465, 4.370, 2.827)
\tkzKiviatLine[thick,color=RadarC6,mark=none,
              fill=RadarC6!10,opacity=.5](4.853, 4.070, 4.960, 3.663)        
\tkzKiviatLine[thick,color=RadarC7,mark=none,
              fill=RadarC7!10,opacity=.5](2.963, 4.270, 1.310, 3.387)              
\tkzKiviatGrad[prefix=,unity=10](1)  
\node[shift={(0.3cm,-0.5cm)}] at (current bounding box.south) 
{
\begin{tabular}{@{}ll@{}}
\ColorBox{RadarC1} mPLUG-owl \hspace*{0.27cm} \ColorBox{RadarC2} LLaMA-AdapterV2 \hspace*{0.27cm} \ColorBox{RadarC7} InstructBLIP\\
\ColorBox{RadarC4} miniGPT-4 \hspace*{0.32cm} \ColorBox{RadarC5} LLaVA \hspace*{0.32cm} \ColorBox{RadarC6} Otter \hspace*{0.32cm} \ColorBox{RadarC3} LaVIN \\
\end{tabular}
};
\end{tikzpicture}
  \vspace{-0.4cm}
  \caption{Zero shot accuracy on each question type.}
  \label{sfig:radar_qtype}
  \end{subfigure}
\caption{Comparison of seven advanced LVLMs on \eval generated MCQs in accuracy (maximum 100).}
\vspace{-0.3cm}
\label{fig:radar_chart}
\end{figure*}

\subsection{Evaluated Models}
We select seven SOTA LVLMs for evaluation and all use the 7B parameters variants for a fair comparison:
\begin{itemize}[noitemsep]
    \item mPLUG-Owl~\cite{ye2023mplugowl}, MiniGPT4~\cite{zhu2023minigpt}, LLaVA~\cite{liu2023LLaVA}, InstructBLIP~\cite{dai2023InstructBlip}: LVLMs based on the visual encoder and the pre-trained LLM that are similar to the architecture we introduce in the \emph{Preliminary Section}. Both follow a two-stage training, where stage-1 is a pre-training stage to align visual and textual concepts when LLM is frozen, and stage-2 is a fine-tuning stage to feed into instruction-following data to train both visual encoder and LLM simultaneously.
    \item Otter~\cite{li2023otter}, LLaMA-AdapterV2~\cite{gao2023adapterv2}, LaVIN~\cite{luo2023LaVIN}: LVLMs with a slightly different architecture, where visual features after adaptation/fusion are attended with text embedding in each/some layers of the LLM component.
\end{itemize}
We also test one large language model (language-only): LLaMA-2-chat~\cite{touvron2023llama2} to verify the necessity of multimodal perception ability for answering the generated MCQs. To extract the choice indices (A, B, C, or D) from the free-form responses of LVLMs, we use regular expressions if they explicitly contain indices. Otherwise, we use ChatGPT to help extract indices.

\subsection{Zero-Shot Setting}
\begin{table*}[h!]
    \centering
    \begin{tabular}{c:cccccc:c}
    \toprule
     & Entity & Number & Color & Material & Action & Spatial & Overall \\
    \midrule
    Human & 94.74 & 83.95 & 75.95 & 79.55 & 91.14 & 70.10 & 81.00 \\
    \hline
    & \multicolumn{7}{c}{\emph{zero-shot}} \\ 
    Random-Guess & 33.33 & 33.33 & 33.33 & 33.33 & 33.33 & 33.33 & 33.33\\
    LLaMA2-chat & 39.20 & 35.13 & 34.60 & 39.60 & 38.07 & 34.67 & 36.88\\
    mPLUG-owl & 40.53 & 34.93 & 32.73 & 33.60 & 34.93 & 30.27 & 34.50\\
    {\footnotesize LLaMA-AdapterV2} & 40.20 & 35.20 & 32.93 & 36.13 & 31.67 & 21.33 & 32.91\\ 
    LaVIN & \underline{47.93} & \textbf{37.27} & 33.33 & \underline{40.47} & 40.46 & 33.53 & 38.83\\
    MiniGPT-4 & 40.00 & 35.07& 33.67 & 35.53 & 38.27 & 34.93 & 36.24\\
    LLaVA & 46.67 & 34.27 & \underline{35.20} & 38.53 & 43.73 & \underline{35.00} & \underline{38.90}\\
    Otter & \textbf{51.13} & \underline{36.87}& \textbf{42.20}& \textbf{44.47} & \underline{46.87} & \textbf{36.13} & \textbf{42.94}\\
    InstructBLIP &30.53& 18.00& 30.40& 37.33& \textbf{49.67}& 31.33& 29.63\\
    \bottomrule
    \end{tabular}
    \caption{Accuracy on \eval different targets of semantic grounding.}
    \label{tab:all_experiments_halluc_types}
\end{table*}

\begin{table}[h!]
    \centering
    \resizebox{0.98\linewidth}{!}{%
    \begin{tabular}{c:cccc:c}
    \toprule
    & YoN & What & FiB & Corr. & Overall \\
    \midrule
    Human & 78.63 & 85.15 & 91.11 &	76.42 &	81.00\\
    \hline
    & \multicolumn{4}{c}{\emph{zero-shot}} \\ 
    {\small Random-Guess} & 50.00 & 25.00 & 25.00 & 25.00 & 33.00\\
    {\small LLaMA2-chat} & 50.20 & 28.55 & 37.20 & 9.00 & 36.88\\
    mPLUG-owl & 48.77 &	25.75 &	35.30 &	25.80 &	34.50\\ 
    {\footnotesize LLaMA-AdapterV2} & \textbf{55.90} & 18.55 &	\textbf{51.90} &	13.17 &	32.91\\ 
    LaVIN & 49.00 &	28.85 &	\underline{51.10} &	31.23 &	38.83\\
    MiniGPT-4 & 47.77 &	30.45 &	37.90 &	28.03 &	36.24 \\
    LLaVA & \underline{50.77} &	34.65 &	43.70 &	28.27 &	\underline{38.90}\\
    Otter & 48.53 & \underline{40.70} &	49.60 &	\textbf{36.63} &	\textbf{42.94} \\
    InstructBLIP & 29.63& \textbf{42.70}&	13.10&	\underline{33.87} & 29.63\\
    \bottomrule
    \end{tabular}%
    }
    \caption{Experimental results (Accuracy) on \eval different types of questions (Q-Type).}
    \label{tab:all_experiments_question_types}
    \vspace{-0.3cm}
\end{table}

\begin{table*}[!ht]
    \centering
    \resizebox{0.97\textwidth}{!}{%
    \begin{tabular}{c:cccccc:r}
    \toprule
     & Entity & Number & Color & Material & Action & Spatial & Overall \\
    \hline
    {\footnotesize LLaMA-AdapterV2+} & 53.00 \underline{\imp{12.80}}	& 47.80 \underline{\imp{12.60}}	& 44.40 \underline{\imp{11.47}}	& 48.93 \imp{12.80}	& 46.20 \imp{14.53}	& \underline{43.53} \textbf{\imp{22.20}}	& 47.31 \underline{\imp{14.40}}\\
    {\small mPLUG-owl+} & 51.61 \imp{11.08} & \underline{48.81} \textbf{\imp{13.48}}  & 39.92 \imp{7.19}  & 43.10 \imp{9.50}        & 44.63 \imp{9.70}   & 39.70 \imp{9.43}        & 43.66 \imp{9.16}\\
    LaVIN+  & 47.20 \imp{9.27} &	\textbf{49.53} \imp{12.26} &	\textbf{66.80} \textbf{\imp{33.47}} &	\textbf{73.33} \textbf{\imp{32.86}} &	\underline{69.99} \textbf{\imp{29.54}} &	\textbf{49.80} \underline{\imp{16.27}} &	\textbf{61.11} \textbf{\imp{22.28}} \\
    LLaVA+ & \underline{60.27} \textbf{\imp{13.60}} & 45.20 \imp{10.93} & 41.27 \imp{6.07} & 60.00 \underline{\imp{21.47}} & \textbf{72.93} \underline{\imp{29.20}} & 37.47 \imp{2.47}  & \underline{52.86} \imp{13.96} \\
    Otter+ & \textbf{62.73} \imp{11.60} & 46.20 \imp{9.33} & \underline{48.60} \imp{6.40} & \underline{61.60} \imp{17.13} & 57.53 \imp{10.66} & 40.47 \imp{4.34} & 52.85 \imp{9.91}\\
    \bottomrule
    \end{tabular}%
    }
    \caption{Accuracy (numbers in regular font) and accuracy gain (numbers in italic font in parentheses) on \eval different grounding targets with enhanced LVLMs (denoted as \emph{[method+]}).}
    \label{tab:instruction_tuned_exp}
\end{table*}   

\begin{table*}[!ht]
\centering
    \begin{tabular}{c:cccc:r}
    \toprule
     & YoN & What & FiB & Corr. & Overall \\
    \hline
    {\footnotesize LLaMA-AdapterV2+} & 51.63 \dtrt{4.27} & 33.75 \underline{\imp{15.20}} & 47.20 \dtrt{4.70} & 52.07 \textbf{\imp{38.90}}& 47.31 \underline{\imp{14.40}}\\
    {\small mPLUG-owl+}  & 56.04 \imp{7.27}	& 37.93 \imp{12.18} &	53.00 \textbf{\imp{17.70}} &	34.72 \imp{8.92} & 43.66 \imp{9.16}\\
    LaVIN+ & \textbf{74.00} \textbf{\imp{25.00}} & \textbf{52.20} \textbf{\imp{23.35}} & 35.90 \dtrt{15.20} & \textbf{62.56} \underline{\imp{31.33}} & \textbf{61.11} \textbf{\imp{22.28}}\\
    LLaVA+ & \underline{61.04} \underline{\imp{10.27}} & \underline{49.65} \imp{15.00} & \textbf{58.30} \underline{\imp{15.60}} & 44.67 \imp{16.40} & \underline{52.86} \imp{13.96}\\
    Otter+ & 53.70 \imp{5.17} & 41.10 \imp{0.40} & \underline{57.90} \imp{8.30} & \underline{57.06} \imp{21.53} & 52.85 \imp{9.91}\\
    \bottomrule
    \end{tabular}%
\caption{Accuracy (numbers in regular font) and accuracy gain (numbers in italic font in parentheses) on \eval different question types with enhanced LVLMs (denoted as \emph{[method+]}).}
\vspace{-0.3cm}
\label{tab:instruction_tuned_exp_q_type}
\end{table*}   

Figure~\ref{fig:radar_chart} provides the radar charts of seven advanced LVLMs on each semantic grounding target or each question type. Among these LVLMs, Otter, LLaVA, and LaVIN are top competitors that consistently deliver better performance than other LVLMs. 
In terms of semantic grounding targets, LVLMs are more effective at perceiving and understanding Entity and Action, but extremely struggle in Spatial relations. We speculate the different performances on different kinds of semantic grounding come from the bias of training data used by LVLMs.
mPLUG-owl, LLaMA-AdapterV2, MiniGPT-4, and InstructBLIP all mainly trained on coarse-grained text-image pairs corpus. On the other hand, Otter has been trained on their curated MIMIC-IT corpus that covers perception, reasoning, and planning-oriented text-image QA pairs. LLaVA has been trained on not only the captions, but also bounding boxes with detailed descriptions. The model checkpoint of LaVIN we used here is trained on ScienceQA~\cite{lu2022scienceQA} corpus, which is in the format of MCQs. These additional training corpora provide more fine-grained information on the multimodal input, thus helping these LVLMs achieve better performance. 

Table~\ref{tab:all_experiments_halluc_types} provides a detailed view of the zero-shot accuracy including human, random-guess, and LLaMA2 baselines. The accuracy of random-guess in each semantic grounding type is always $33.33$, since each semantic grounding subset contains 500 two-way ($50.00$ accuracy by random guess) and 1000 four-way MCQs ($25.00$ accuracy by random guess). Human performance (five annotators carefully answer random 500 MCQs from \eval) sets up the upper bound of the zero-shot experiments, which significantly outperforms all other LVLMs. Interestingly, humans do not perform well in color, material, and spatial. 
One possible reason for human deficiency in color and material is that human perception organs are a bit weaker at distinguishing attributes, as compared to recognizing objects and actions.
For Spatial MCQs, we observe that the source data quality is not high. For example, sometimes the spatial relations are labeled according to the physical position, while sometimes they are labeled according to what the annotator perceived.
Comparing the language only LLaMA2-chat and other LVLMs, the best LVLMs are consistently better in every semantic grounding type. This indicates the importance of the perception ability of input vision modality.
Besides, LVLM models struggle more with correction questions, which is also more challenging for humans.

Moreover, Table~\ref{tab:all_experiments_question_types} supplies LVLMs performance on different Q-Types with human, random-guess, and LLaMA2 baselines. As can be seen, Correction type MCQs are most challenging for both humans and LVLMs. Given the accuracy of the Random-Guess baseline as $25.00$ (four-way MCQs with one correct answer), the best LVLM (Otter) only outperforms it by $11.63$. While for other four-way What and Fill-in-the-Blank MCQs, the best LVLM achieves $17.70$ and $26.10$ accuracy gains, separately. On the other hand, Yes-or-No type MCQs are quite challenging too. Given the Random-Guess accuracy as $50.00$, the best LVLM (LLaMA-Adapter) only achieves $5.9$ higher accuracy. 
The patterns of experimental outcomes indicate that LVLMs are better at responding to descriptive queries (What and Fill-in-the-Blank MCQs). One speculation for that is these LVLMs have been trained on a great amount of image captioning data and image description instruction data. With the pretraining on such training corpora, LVLMs may be relatively less prepared for judgemental and corrective queries.

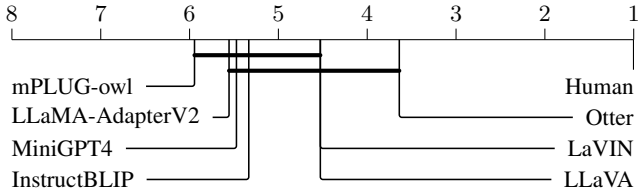
\begin{figure}[!htb]
\begin{tikzpicture}[scale=0.98]
\begin{axis}[
  axis x line=center, axis y line=none, xmin=1, xmax=8, ymin=-5.5, ymax=0, scale only axis, height=6\baselineskip, width=\axisdefaultwidth, ticklabel style={anchor=south, yshift=1.33*\pgfkeysvalueof{/pgfplots/major tick length}, font=\small}, every tick/.style={yshift=.5*\pgfkeysvalueof{/pgfplots/major tick length}}, axis line style={-}, title style={yshift=\baselineskip}, x dir=reverse
]

\draw[semithick, rounded corners=1pt] (axis cs:1.0, 0) |- (axis cs:1.0, -1.5) node[font=\small, fill=white, inner xsep=5pt, outer xsep=-5pt, anchor=east] {Human};

\draw[semithick, rounded corners=1pt] (axis cs:3.638888888888889, 0) |- (axis cs:1.0, -2.5) node[font=\small, fill=white, inner xsep=5pt, outer xsep=-5pt, anchor=east] {Otter};

\draw[semithick, rounded corners=1pt] (axis cs:4.527777777777778, 0) |- (axis cs:1.0, -3.5) node[font=\small, fill=white, inner xsep=5pt, outer xsep=-5pt, anchor=east] {LaVIN};

\draw[semithick, rounded corners=1pt] (axis cs:4.527777777777778, 0) |- (axis cs:1.0, -4.5) node[font=\small, fill=white, inner xsep=5pt, outer xsep=-5pt, anchor=east] {LLaVA};

\draw[semithick, rounded corners=1pt] (axis cs:5.333333333333333, 0) |- (axis cs:8.0, -4.5) node[font=\small, fill=white, inner xsep=5pt, outer xsep=-5pt, anchor=west] {InstructBLIP};

\draw[semithick, rounded corners=1pt] (axis cs:5.472222222222222, 0) |- (axis cs:8.0, -3.5) node[font=\small, fill=white, inner xsep=5pt, outer xsep=-5pt, anchor=west] {MiniGPT4};

\draw[semithick, rounded corners=1pt] (axis cs:5.555555555555555, 0) |- (axis cs:8.0, -2.5) node[font=\small, fill=white, inner xsep=5pt, outer xsep=-5pt, anchor=west] {LLaMA-AdapterV2};

\draw[semithick, rounded corners=1pt] (axis cs:5.944444444444445, 0) |- (axis cs:8.0, -1.5) node[font=\small, fill=white, inner xsep=5pt, outer xsep=-5pt, anchor=west] {mPLUG-owl};

\draw[ultra thick, line cap=round] (axis cs:4.527777777777778, -0.5) -- (axis cs:5.944444444444445, -0.5);

\draw[ultra thick, line cap=round] (axis cs:3.638888888888889, -1.0) -- (axis cs:5.555555555555555, -1.0);
\end{axis}
\end{tikzpicture}
\vspace{-0.2cm}
\caption{The critical difference diagrams of each LVLM for each semantic grounding target with each question type. The lower rank (further to the right) represents the better performance. LVLMs connected by thick bars indicate that these models are not significantly different ($p < 0.05$).}
\label{fig:CD_diagram}
\vspace{-0.3cm}
\end{figure}

We further conduct an in-depth analysis using critical difference (CD) analysis~\cite{demsar2006statistical}. Figure~\ref{fig:CD_diagram} shows the CD diagram, which is a powerful tool to compare outcomes of multiple compared models over multiple observations. The CD analysis involves several hypothesis tests, and models connected with each other in the diagram means that the performances of these models are not that different in the sense of statistical significance. As shown in Figure~\ref{fig:CD_diagram}, the performances of LVLMs are relatively similar, although they can be divided into two groups. Noticeably, there still exists a significant gap between humans and all LVLMs.
Overall, \eval serves as a good evaluation module for researchers to understand the limitations of LVLMs with regard to specific fine-grained semantic grounding.

\subsection{Enhanced LVLMs Setting}
In order to enhance semantic grounding in LVLMs, we conduct instruction tuning\footnote{Instruction tuning details are elaborated in the Appendix.} on LVLMs by generating 180K fine-grained multimodal instruction data covering multi-round conversations, vision-prompted recognition, and fact-checking. Table~\ref{tab:instruction_tuned_exp} shows the performance gains on several enhanced LVLMs by instruction tuning. Following the same format, we use \textbf{bold font} to highlight the best accuracy or accuracy gain, and \underline{underline font} to highlight the second-best accuracy or accuracy gain. For those LVLMs not included, they typically do not release well-established instruction tuning scripts (or scripts are specifically for image captioning tasks). As can be seen, we observe consistent improvements over all tuned LVLMs, which indicates the effectiveness of our enhancement method. Interestingly, the performance gain for LAVIN is significantly higher than for other models. We believe this is primarily due to LAVIN incorporating trainable adaptors into both the vision encoder and the LLM, enabling end-to-end optimization of the entire model.
Moreover, we supply Table~\ref{tab:instruction_tuned_exp_q_type} that offers a detailed breakdown of accuracy improvements based on MCQ types. A closer investigation of the table reveals that while performance enhancements are evident across most MCQs, there are instances of decreased accuracy in specific MCQ types. For instance, LaVIN exhibits a notable decrease of $\downarrow 15.20$ accuracy in Fill-in-the-Blank MCQs, while LLaMA-AdapterV2 records slightly lower accuracies in Yes-or-No and Fill-in-the-Blank MCQs. Despite these isolated variations, consistent performance improvements are observed in other categories. 
In summary, LVLMs enhanced with our automatically generated instruction data deliver better semantic grounding performance. The instruction data is fundamentally different from the training data, since instruction data does not share the same data distribution and formats of the \eval testing data.
The instruction data also serves as a valuable resource for instruction tuning, readily available for any LVLMs to utilize.
\section{Conclusion and Future Work}
In this work, we evaluate and enhance the ability of LVLMs to ground fine-grained vision-language inputs. We propose an evaluation method \eval to automatically generate testing samples focusing on specific grounding targets in various question formats, along with comprehensive experiments to understand how well the current state-of-the-art LVLMs perform in semantic grounding. We further propose a data-centric approach to enhance LVLMs. Its effectiveness is validated by observing consistent performance improvement of these LVLMs, after instruction tuning on multimodal multil-round conversations generated by the enhancement method.
In the future, we aim to extend the scope of semantic grounding in LVLMs into (1) more modalities such as audio, time-series, and tabular; (2) more semantic grounding targets such as social-emotion, terrains, and human organs.

\bibliography{aaai24}

\appendix
\clearpage
\appendixpage
\begin{appendices}

\section{Implementations of Compared LVLMs}
Implementations of the compared baselines are from codebases of original authors, with their open-sourced URLs as follows:
\begin{itemize}
    \item LLaMA2-chat~\cite{touvron2023llama2}: \url{https://github.com/facebookresearch/llama}, with the \emph{LLaMA2-7B-chat} checkpoint.
    \item mPLUG-owl~\cite{ye2023mplugowl}: \url{https://github.com/X-PLUG/mPLUG-Owl}, with the \emph{mPLUG-Owl-7B} checkpoint.
    \item LLaMA-AdapterV2~\cite{gao2023adapterv2}: \url{https://github.com/OpenGVLab/LLaMA-Adapter}, with the \emph{LLaMA-AdapterV2-multimodal} checkpoint.
    \item LaVIN~\cite{luo2023LaVIN}: \url{https://github.com/luogen1996/LaVIN}, with the \emph{LaVIN-7B} checkpoint.
    \item MiniGPT-4~\cite{zhu2023minigpt}: \url{https://github.com/Vision-CAIR/MiniGPT-4}, with the \emph{MiniGPT-4-Vicuna7B-V0} checkpoint.
    \item LLaVA~\cite{liu2023LLaVA}: \url{https://github.com/haotian-liu/LLaVA}, with the \emph{LLaVA-Vicuna7B-v1.1} checkpoint.
    \item Otter~\cite{li2023otter}: \url{https://github.com/Luodian/Otter}, with the \emph{OTTER-Image-MPT7B} checkpoint.
    \item InstructBLIP~\cite{dai2023InstructBlip}: \url{https://github.com/salesforce/LAVIS/tree/main/projects/instructblip}, with the \emph{blip2-vicuna7B-instruct} checkpoint.
\end{itemize}
As can be seen, we use 7B parameter size checkpoints for all compared LVLMs and the text-only LLaMA2-chat. All LVLMs utilize LLaMA1 as the large language model component for a fair comparison.
The hyper-parameters for inference (\textit{e.g.} temperature, number of beams, \textit{etc.}) for each LVLM are directly taken from the default ones recommended by their authors in the original codebases, respectively. 

For instruction tuning, we also keep the default hyper-parameters (\textit{e.g.} batch size, optimizer, learning rate, \textit{etc.}) except explicitly set the \emph{max tuning epoch} as $3$ for all LVLMs.
Specifically, for instruction tuning of LLaMA-AdapterV2, the visual projector and language model adaptor are updated while leaving the pre-trained visual encoder and the pre-trained language model frozen.
For mPLUG-Owl, only the pre-trained language model is updated.
For LaVIN, the visual encoder adaptors, the visual projector, and the language model adaptors (namely ``Mixture-of-Modality Adapter'' in the original paper) are updated.
For LLaVA, the visual projector and the language model are updated.
For Otter, the Perceiver resampler of the visual encoder, input/output embeddings of the language model, and the cross-attention layers are updated.

\begin{table*}[tp]
\centering
\begin{tabular}{lp{0.8\linewidth}}
\toprule
\textbf{Case\#1} \\
& \includegraphics[width=0.5\linewidth]{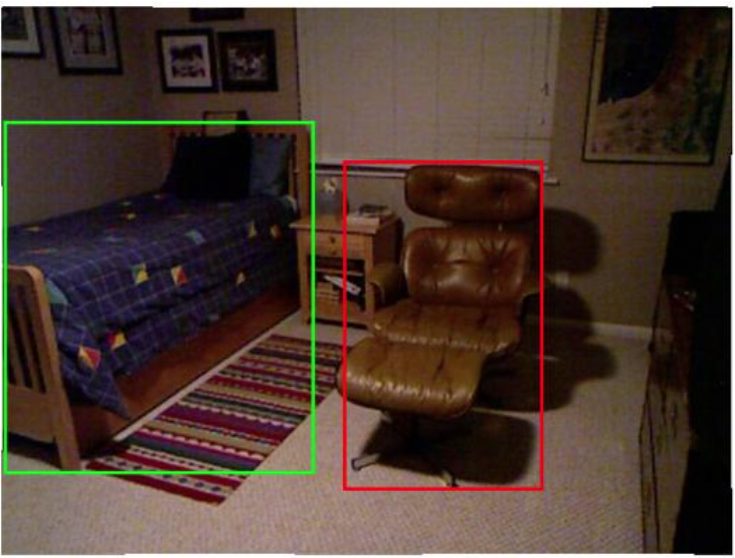}\\
& {\footnotesize Source: SpatialSense~\cite{yang2019spatialsense}}\\
User & What is in the green bounding box in the image?\\
LLaVA & In the green bounding box in the image, there is \textcolor{orange}{a chair.}\\
LLaVA+ & The image shows \textcolor{cyan}{the bed} located in the green bounding box.\\
\hline
User & What is in the red bounding box in the image?\\
LLaVA & In the red bounding box in the image, there is \textcolor{orange}{a bed.}\\
LLaVA+ & The image shows \textcolor{cyan}{the chair} located in the green bounding box.\\
\hline
User & What is the spatial relationship between them? \\
LLaVA & The bed is located next to the chair, with the chair positioned \textcolor{orange}{in front of} the bed.\\
LLaVA+ & In the image, the bed in the green bounding box is \textcolor{cyan}{to the left} of the chair in the red bounding box.\\
\toprule
\textbf{Case\#2} \\
& \includegraphics[width=0.5\linewidth]{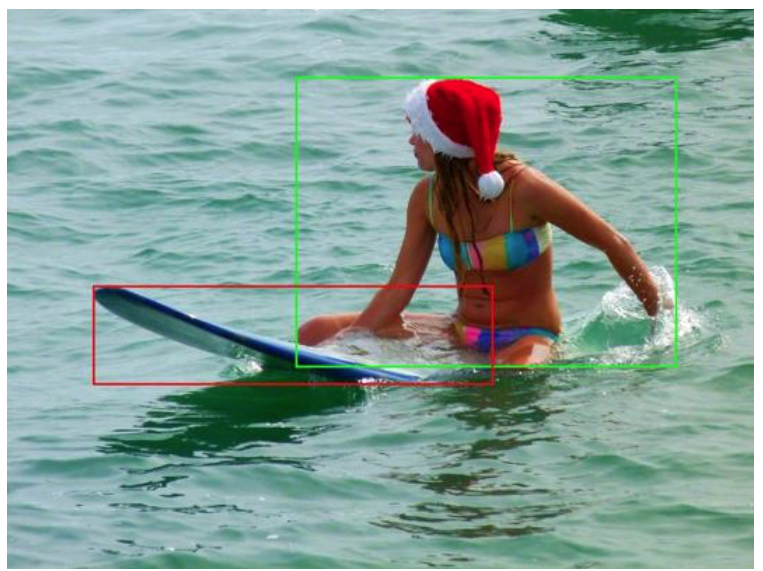}\\
& {\footnotesize Source: OpenImage-V7~\cite{OpenImages}}\\
User & What is in the green bounding box?\\
LLaVA &  In the green bounding box in the image, there is \textcolor{cyan}{a woman wearing a Santa hat} while riding a surfboard in the ocean.\\
LLaVA+ & The image displays \textcolor{cyan}{the woman} located in the green bounding box. \\
\hline
User & What is in the red bounding box?\\
LLaVA & In the red bounding box, there is a \textcolor{orange}{close-up of a woman} wearing a Santa hat while riding a surfboard in the ocean.\\
LLaVA+ & The \textcolor{cyan}{surfboard is situated} within the red bounding box in the picture.\\
\hline
User & What is the interaction between them?\\
LLaVA & The interaction between the objects in the red bounding box and the green bounding box is that \textcolor{orange}{the woman in the red box is riding a surfboard} in the ocean, while the woman in the green box is also riding a surfboard in the ocean.\\
LLaVA+ & In the context of the image, the woman in the red bounding box \textcolor{cyan}{surf} the surfboard in the green bounding box. \\
\bottomrule
\end{tabular}
\caption{Case study regarding spatial misgrounding. LLaVA+ denotes the enhanced LLaVA using our proposed instruction tuning data.\textcolor{cyan}{Text in cyan} indicates the precise statement. \textcolor{orange}{Text in orange} indicates the ambiguous statement.}
\label{tab:case_study}
\end{table*}

\section{Case Studies for Enhanced LVLMs}
We investigate the differences between responses from original LVLMs and responses from treated LVLMs. Table~\ref{tab:case_study} shows two real examples using the same data source with free-form questions. The compared LVLM is LLaVA, between the original 7B checkpoint version and the instruction-tuned version (denoted by \emph{[method+]}). 
In Case \#1, the image depicts a well-furnished bedroom softly illuminated, creating a cozy and inviting ambiance. A red bounding box highlights a chair, while a green bounding box highlights a bed adorned with a blue comforter and pillows. Before instruction tuning, LLaVA struggled to provide accurate responses to questions about this image. For example, LLaVA mistakenly identified the chair as highlighted by the green bounding box, and the bed as highlighted by the red bounding box. In contrast, LLaVA+ produced more accurate responses to these questions. Furthermore, when asked about the spatial relationship between the chair and the bed, LLaVA responded incorrectly, stating that ``the chair is in front of the bed.'' In contrast, LLaVA+ correctly indicated that the bed is to the left of the chair.
In Case\#2, the image features a woman wearing a Santa hat while riding a surfboard in the water. This time, both LLaVA and LLaVA+ successfully identify the object within the green bounding box, which is the woman. When moving to the red bounding box, LLaVA fails to ground it and still recognizes it as the woman, while LLaVA+ adeptly recognizes it as the surfboard.  LLaVA's failure with the red bounding box results in an incorrect response to the subsequent question concerning the interaction between these two bounding boxes. 
These two illustrative examples support the effectiveness of our proposed data-centric enhancement method in aiding LVLMs in recognizing and interpreting nuanced multimodal information.

\end{appendices}

\end{document}